\documentclass[lettersize,journal]{IEEEtran}
\usepackage{amsmath,amsfonts,amssymb}
\usepackage{algorithm}
\usepackage{algorithmic}
\usepackage{array,booktabs,multirow}
\usepackage{graphicx}
\usepackage{textcomp}
\usepackage{cite}
\usepackage{url}
\usepackage{placeins}
\usepackage{stfloats}
\usepackage[T1]{fontenc}
\graphicspath{
    {./fig/}
}
\def\BibTeX{{\rm B\kern-.05em{\sc i\kern-.025em b}\kern-.08em
    T\kern-.1667em\lower.7ex\hbox{E}\kern-.125emX}}
\begin{document}

\title{Direct Action-Manifold Modeling for Efficient and Robust Bimanual Manipulation}

\author{
Hongren Wang,
Mingyu You,~\IEEEmembership{Member, IEEE},
Zihan Li,
Zhipeng Wang,
Shuo Jiang,~\IEEEmembership{Member, IEEE},
and Bin~He,~\IEEEmembership{Senior Member, IEEE}%
\thanks{This work was supported by the National Key R\&D Program of China under Grant 2024YFB4707600, in part by the National Natural Science Foundation of China under Grants 62688201, 62573321, and W2441018, and in part by the Shanghai Municipal Commission of Science and Technology under Grant 24511104400. (\emph{Corresponding author: Bin He.})}%
\thanks{Hongren Wang is with the Shanghai Research Institute for Intelligent Autonomous Systems, Tongji University, Shanghai 201210, China (e-mail: hongrenwang@tongji.edu.cn).

Zihan Li is with the College of Electronics and Information Engineering, Tongji University, Shanghai 201804, China (e-mail: zihan\_li@tongji.edu.cn).

Mingyu You, Zhipeng Wang, Shuo Jiang, and Bin He are with the College of Electronics and Information Engineering, Tongji University, Shanghai 201804, China, and also with the Shanghai Research Institute for Intelligent Autonomous Systems, Tongji University, Shanghai 201210, China (e-mail: myyou@tongji.edu.cn; wangzhipeng@tongji.edu.cn; jiangshuo@tongji.edu.cn; hebin@tongji.edu.cn).}%
}
\maketitle

\begin{abstract}
Diffusion and flow matching visuomotor policies require iterative denoising or numerical integration for multimodal action generation, introducing latency and execution mismatch. We instead consider a different formulation: conditioned on vision and proprioception, demonstrated bimanual action chunks concentrate in a structured observation-dependent region of joint-action space, making the action target more direct to model than the generative process. Based on this view, we propose the Direct Manifold Action Transformer (DMAT), which reformulates action generation as the direct modeling of clean, observation-conditioned joint-space action chunks from stochastic interpolants, thereby enabling single-step inference without relying on iterative generative inference or auxiliary trajectory-straightening procedures. Within DMAT, local 3D geometry anchors predictions to spatial structure, while State-Consistent Spectral-Time Refinement (STR) reconciles predicted chunks with proprioception and suppresses high-frequency residuals. Across 567 standard and out-of-distribution (OOD) trials per method, DMAT outperformed Improved 3D Diffusion Policy (IDP3) and DiTFlow3D in aggregate success rate, achieving relative improvements of 129.1\% and 95.7\%, respectively, while generating action chunks 2.06$\times$ and 1.34$\times$ faster and remaining robust under repeated 50 N interventions. These results pave the way for a novel generative paradigm for efficient and robust bimanual manipulation.
\end{abstract}


\section{Introduction}\label{sec:introduction}
\IEEEPARstart{G}{enerative} visuomotor policies support flexible robotic automation by connecting visual and proprioceptive information to coordinated actions. Bimanual coordination, sustained contact, and material transfer require actions that remain coherent as visual and proprioceptive states evolve \cite{zhao2023learning}, \cite{kim2024goal}. The information-processing challenge spans both prediction and delivery: spatial observations must inform temporally coordinated action chunks, and those chunks must remain compatible with the reached robot state.

Diffusion and flow matching policies construct expressive actions through a learned generative process. Diffusion policies iteratively denoise action sequences \cite{chi2025diffusion}, whereas flow matching policies integrate learned velocity fields \cite{lipman2022flow}, \cite{ding2025fast}. Consistency, distillation, and few-step methods reduce this cost \cite{prasad2024consistency}, \cite{zhang2025flowpolicy}, while the remaining policy computation can delay the next command chunk \cite{black2025real}.

\begin{figure}[!t]
\centering
\includegraphics[width=0.95\columnwidth]{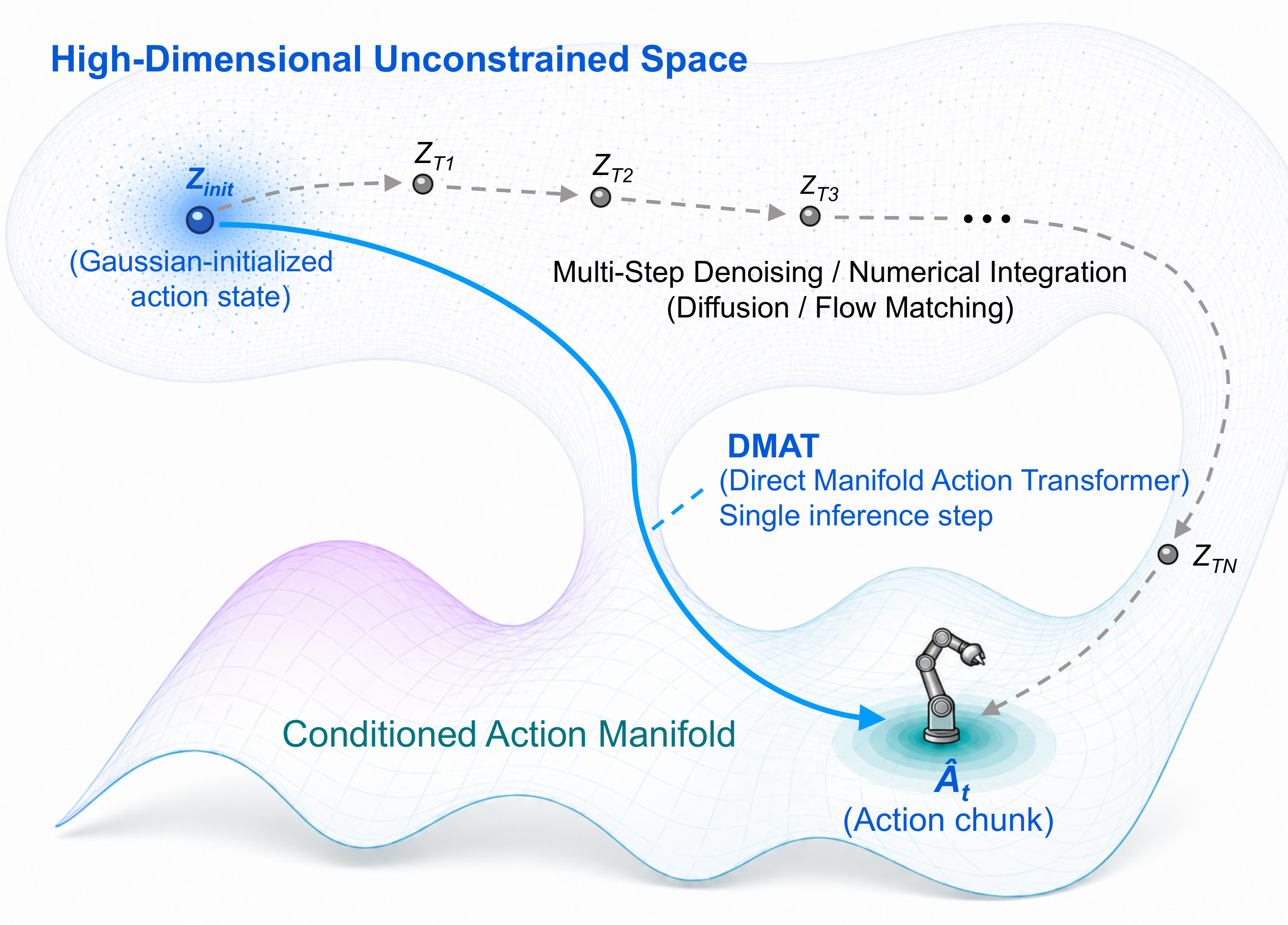}
\caption{Conceptual comparison of action generation formulations. Representative multistep diffusion and flow matching policies evolve a Gaussian-initialized action state $Z_{\mathrm{init}}$ through intermediate states $Z_{\tau_i}$. DMAT directly predicts the clean observation-conditioned chunk $\widehat A_t$ in one evaluation. The lower region illustrates hypothesized observation-conditioned structure in demonstrated actions.}
\label{fig:principle-comparison}
\end{figure}

The central question is therefore whether a generative path should remain the primary modeling target for a robot action chunk. Joint commands are lower-dimensional than images and constrained by configuration, coordination, local geometry, and task phase; under a fixed observation, demonstrations can occupy a structured region of joint-action space. We use this property to define direct clean action prediction from a stochastic interpolant, rather than incrementally shortening an existing denoising or integration procedure. This is a change in modeling viewpoint, not merely in inference step count (Fig.~\ref{fig:principle-comparison}).

Direct chunk prediction must retain spatial information for grasping, contact, and coordinated transport. Pooled pointcloud representations supply a shared scene summary \cite{ze20243d}, \cite{ze2025generalizable}, whereas DMAT gives each temporal action token access to unpooled 3D features.

Geometrically grounded prediction alone does not ensure robust action chunk execution. Servo error, contact, or intervention can move the robot away from the state assumed by queued references. Real-Time Chunking (RTC) corrects latency-induced observation--action misalignment \cite{black2025real}, but does not explicitly reconcile a new action chunk with physical state deviations. Robust execution therefore requires state consistency at regeneration boundaries.

These requirements motivate the \emph{conditional action-manifold hypothesis}: observation-conditioned demonstrations concentrate around structured bimanual action patterns shaped by configuration, local geometry, coordination, and task phase. DMAT operationalizes this hypothesis with clean action targets, local geometric conditioning, and state-consistent delivery.

We introduce \emph{direct action-manifold modeling} and instantiate it with the DMAT. Under the conditional action-manifold hypothesis, DMAT couples single-step clean action prediction with local geometric conditioning and STR for physical execution (Fig.~\ref{fig:framework}). The contributions are:
\begin{enumerate}
\item We formulate bimanual action generation around the conditional action-manifold hypothesis and make clean observation-conditioned action chunks the primary modeling target. By construction, DMAT maps stochastic interpolants directly to these action chunks without solving an iterative denoising chain or numerically integrating a learned generative velocity field.
\item We develop a local geometry-anchored action space decoder that preserves position-specific 3D conditioning. Temporal action tokens query unpooled spatial tokens through cross-attention, retaining the coupling between action sequences and scene structure.
\item We introduce STR to carry directly generated chunks into state-consistent action chunk execution. It reconnects pending actions to the measured joint state and attenuates high-frequency residuals at regeneration boundaries.
\item We validate the complete chain on a Fourier GR2 robot through standard, OOD, and prescribed 50 N lateral-pull tests. Progressive ablations evaluate successive additions of geometric conditioning and STR to the direct predictor.
\end{enumerate}

\section{Related Work}\label{sec:related-work}

\subsection{Visuomotor Generative Policies}
Generative visuomotor policies combine action modeling with task-relevant conditioning. ARViP combines adversarial learning with joint-level temporal regularization \cite{wang2026arvip}. Diffusion Policy iteratively denoises actions \cite{chi2025diffusion}; subsequent variants incorporate stage-dependent visual conditioning \cite{zhao2026staged}. Double Diffusion Policy uses human-action guidance to reduce dependence on expert demonstrations \cite{liang2026double}. Conditional flow matching learns generative velocity fields \cite{lipman2022flow}. Flow Policy uses tailored flow trajectories and self-supervised representations \cite{fang2025flow}, while Riemannian variants respect geometric action spaces \cite{ding2025fast}.

Recent one-step and few-step policies reduce online generation cost through consistency learning, distillation, or shortened integration \cite{prasad2024consistency}, 
\cite{zhang2025flowpolicy}, \cite{yan2025maniflow}, while retaining the generative path as the organizing object. In human motion prediction, MotionFlow learns average velocities for one- or few-step generation \cite{zhong2026efficient}. DMAT instead models the clean action directly from an interpolated state, making a formulation-level change in the modeling target rather than another schedule for traversing the same path.

\subsection{Spatial Geometric Representations in Manipulation}
Spatial representations range from scene summaries to local geometric features. MSP-Grasp combines global, regional, and local perception for cluttered grasping \cite{liu2026mspgrasp}. PerAct and RVT use voxel or multiview representations \cite{shridhar2023perceiver}, \cite{goyal2023rvt}; ACT3D and RISE use 3D features for action selection \cite{gervet2023act3d}, \cite{wang2024rise}; and 3D Diffusion Policy (DP3) and IDP3 condition policies on sparse pointclouds \cite{ze20243d}, \cite{ze2025generalizable}. PointNet-style pooling is efficient \cite{qi2017pointnet}, whereas hierarchical models retain neighborhoods before aggregation \cite{qi2017pointnet++}. One pooled vector, however, gives every future position the same scene summary.

Surface-constrained policies encode task geometry explicitly before generating executable motion \cite{ke2026surface}. DMAT retains unpooled spatial tokens as cross-attention memory, conditioning different portions of a joint space chunk on local scene features.

\subsection{Action Chunk Execution Under Physical State Deviations}
Action chunk execution methods preserve continuity within predictions or across replanning events. Temporal ensembling blends overlaps \cite{zhao2023learning}; optimization constrains trajectories \cite{xu2026diffog}, \cite{xu2024leto}; and RTC corrects latency-induced misalignment \cite{black2025real}. Queued references can nevertheless disagree with measured proprioception after contact, servo error, or intervention.

Execution robustness also depends on recovering from states outside demonstration coverage. Phase-conditioned policies add force-aware recovery \cite{chen2026phaseconditioned}. HReDiff improves continual diffusion imitation through hierarchical replay, but reports recovery failures after accumulated trajectory deviations \cite{wei2026hierarchically}. SCP maps diffusion-generated intent to surface-constrained dynamic movement primitives \cite{ke2026surface}. STR instead reconnects a pending joint-space chunk to the measured state and shapes its temporal residuals before delivery.

\section{Methods}\label{sec:method}
\begin{figure*}[!t]
\centering
\includegraphics[width=0.95\textwidth]{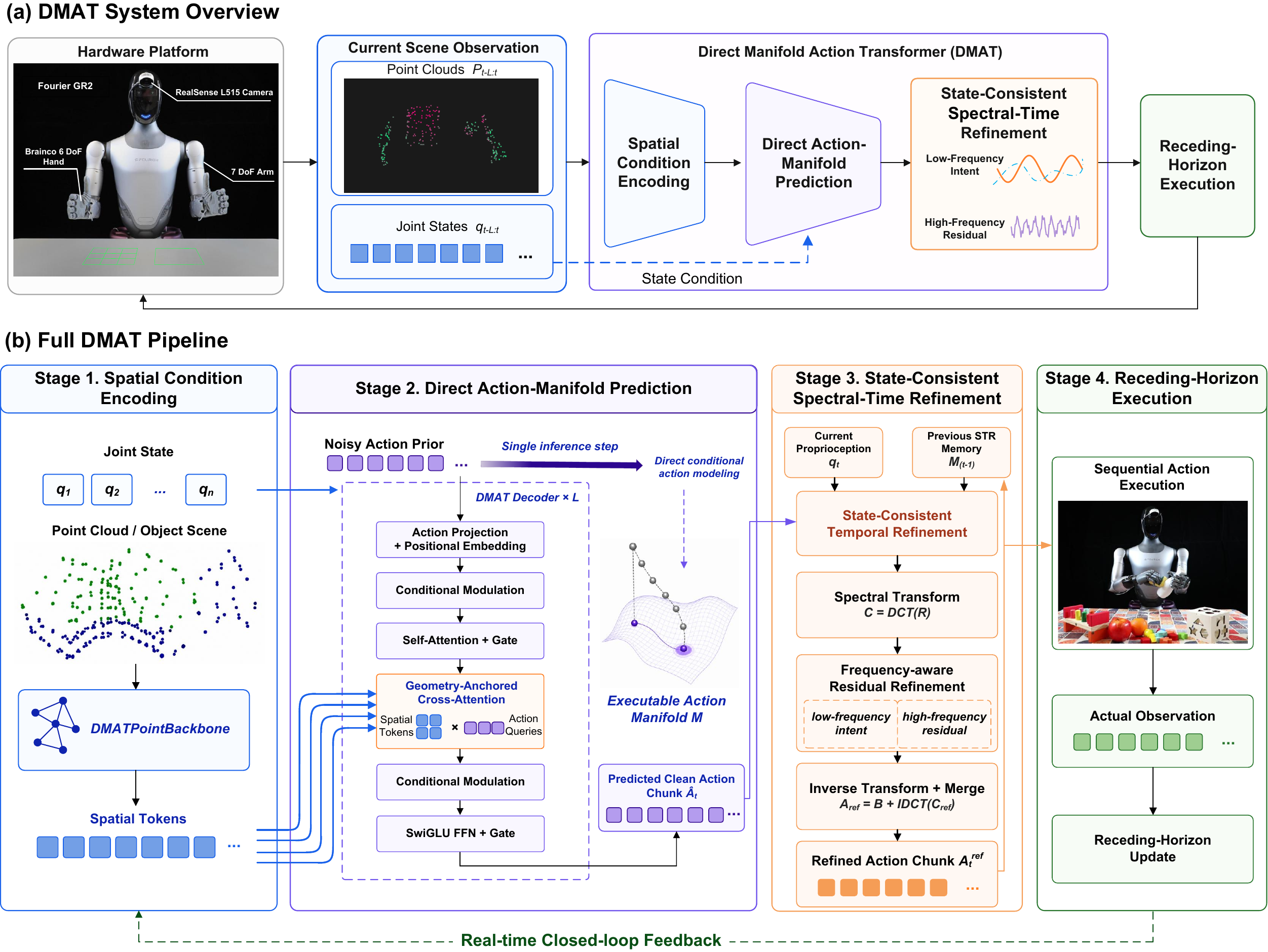}
\caption{DMAT system and pipeline. (a) System overview integrating direct action-manifold prediction, STR, and receding-horizon execution. (b) Detailed pipeline. Stage~1 encodes recent pointclouds into spatial tokens; Stage~2 directly predicts the clean observation-conditioned action chunk $\widehat{A}_t$ from a Gaussian-initialized action state and spatial-proprioceptive conditions; Stage~3 combines $\widehat{A}_t$, the measured joint state, and STR memory to preserve low-frequency intent while attenuating high-frequency residuals; and Stage~4 executes $A_t^{\mathrm{STR}}$ in a receding horizon and returns updated observations.}

\label{fig:framework}
\end{figure*}

DMAT combines direct clean-action prediction, local 3D conditioning, and state-consistent action chunk execution in a receding-horizon perception-to-execution chain (Fig.~\ref{fig:framework}).

\subsection{Task Definition and Joint-Space Action Representation}\label{sec:task-definition}
\subsubsection{Observation and Action Spaces}
We formulate bimanual manipulation as observation-conditioned joint control through action chunks learned from teleoperated demonstrations. At time $t$, the policy observation is
\begin{equation}\label{eq:observation}
o_t=\left(P_{t-K+1:t},\mathbf q_{t-K+1:t}\right),
\end{equation}
where $P_{t-K+1:t}$ and $\mathbf q_{t-K+1:t}$ are the most recent $K$ pointclouds and joint states. The joint-state and action dimensions are $d_q$ and $d_a$, and $\mathbf a_t$ is the joint-position target sent to the servo.

\subsubsection{Action Chunk Prediction}
The policy output aligns the observation history with a complete joint-space action chunk
\begin{equation}\label{eq:action-chunk}
A_t=\left[\mathbf a_{t+1-K}^\top;\ldots;\mathbf a_{t+H-K}^\top\right]
\in\mathbb{R}^{H\times d_a}.
\end{equation}
With $K=2$, the first row is history aligned and is not executed. The executable suffix is $A_t^{\mathrm{exe}}=A_t[s_e{:}H]\in\mathbb R^{H_s\times d_a}$, where $s_e=K$ and $H_s=H-s_e+1$. The policy queues $H_e\leq H_s$ targets and regenerates a chunk when the queue is empty.

A unified joint-space vector coordinates both arms, both hands, the head, and the fixed waist channel. States and action targets use separate min--max statistics; STR operates on normalized representations before targets are restored for servoing.

\subsection{Direct Conditional Action Generation Under the Proposed Conditional Action-Manifold Hypothesis}\label{sec:direct-generation}
\subsubsection{Conditional Action-Manifold Hypothesis}
The direct formulation assumes that observation-conditioned demonstrations concentrate in a structured region of executable action space. Robot configuration, joint range, coordination, local geometry, and task progress constrain $p_{\mathcal D}(A_t^{\mathrm{exe}}\mid o_t)$; we call this task-dependent structure the conditional action manifold and use it to motivate a direct clean action target \cite{loaiza2024deep}.

\subsubsection{Mathematical Formulation of Direct Action Mapping}
DMAT realizes this hypothesis through clean action regression at randomly sampled interpolation levels. During training, it interpolates between Gaussian action states and demonstrations while predicting the clean chunk at every level:
\begin{equation}\label{eq:training-sampling}
(o_t,A_t)\sim\mathcal D,\;
\operatorname{vec}(E)\sim
\mathcal N(\mathbf 0,\sigma^2\mathbf I_{Hd_a}),\;
\tau\sim\mathcal U(0,1),
\end{equation}
where $E\in\mathbb R^{H\times d_a}$ has independent Gaussian entries, $\sigma$ is the noise scale, and $\tau$ is sampled independently of $(o_t,A_t)$ and $E$. The interpolated state is
\begin{equation}\label{eq:interpolation}
Z_\tau=(1-\tau)E+\tau A_t.
\end{equation}
The clean prediction is
\begin{equation}\label{eq:clean-prediction}
\widehat A_t=f_\theta(Z_\tau,\tau,o_t).
\end{equation}
The clean action regression objective is
\begin{equation}\label{eq:dmat-loss}
\begin{aligned}
\mathcal L_{\mathrm{DMAT}}
&=\mathbb E_{\substack{(o_t,A_t)\sim\mathcal D,\,E,\,\tau}}
\left[\frac{\left\|f_\theta(Z_\tau,\tau,o_t)-A_t\right\|_F^2}{Hd_a}\right],
\end{aligned}
\end{equation}
where $\|\cdot\|_F$ is the Frobenius norm.

\subsubsection{Single Inference Step Generation and Execution}
At deployment, DMAT evaluates the trained clean action map once rather than traversing the interpolation path:
\begin{equation}\label{eq:inference}
\begin{aligned}
\operatorname{vec}(Z_{\mathrm{init}})&\sim\mathcal N(\mathbf 0,\sigma^2\mathbf I_{Hd_a}),\\
\widehat A_t&=f_\theta(Z_{\mathrm{init}},0,o_t).
\end{aligned}
\end{equation}
The action chunk is therefore generated in a single inference step through one network evaluation, without iterative denoising or online velocity-field integration. We assess task success, generation latency, and measured motion during physical execution.

\subsubsection{Difference From Existing Generative Policies}
DMAT trains a clean-action predictor across randomly sampled interpolation levels and evaluates it once at the pure-noise endpoint. Unlike observation-only action regression, training also conditions on stochastic interpolants. The experiments evaluate this formulation together with local geometric conditioning and STR.

\subsection{Local Geometry-Anchored Action Space Decoder}\label{sec:spatial-decoder}
\subsubsection{Spatial Token Encoding}
To ground a generated chunk in the scene, the decoder preserves local 3D conditions for every temporal position. Each point is mapped to a coordinate feature and positional embedding, and spatial self-attention produces unpooled memory $S\in\mathbb R^{N_s\times d}$. Recent frames are concatenated without explicit identities or temporal correspondence, leaving each action position access to local tokens.

\subsubsection{Conditional Action Tokens}
Conditional action tokens separate temporal coordination, local spatial evidence, and proprioceptive state. The decoder forms $X^{(0)}=\phi_a(Z_\tau)+E_{\mathrm{act}}$ and $\mathbf c=\phi_c([\phi_\tau(\tau);\phi_q(\mathbf q_{t-K+1:t})])$. Adaptive normalization injects $\mathbf c$, while gated self-attention coordinates the $H$ positions.

\subsubsection{Action Space Decoding}
Spatial cross-attention lets each temporal action token form its own condition over the unpooled spatial token memory. We use root-mean-square normalization (RMSNorm) before constructing the cross-attention queries. For head $h$, define $Q_h=\operatorname{RMSNorm}_{\mathrm{ca}}^{(\ell)}(X_{\mathrm{sa}}^{(\ell)})W_{Q,h}^{(\ell)}$, $K_h=SW_{K,h}^{(\ell)}$, and $V_h=SW_{V,h}^{(\ell)}$, with per-head dimension $d_h$ and $Md_h=d$. The spatial cross-attention update is
\begin{equation}\label{eq:decoder-cross-attention}
\begin{aligned}
O_h
&=\operatorname{Softmax}_{\mathrm{row}}\!\left(
\frac{Q_hK_h^\top}{\sqrt{d_h}}\right)V_h,\\
X_{\mathrm{ca}}^{(\ell)}&=X_{\mathrm{sa}}^{(\ell)}+
\operatorname{Concat}(O_1,\ldots,O_M)W_O^{(\ell)}.
\end{aligned}
\end{equation}
Rowwise softmax preserves position-specific weights over spatial tokens. A condition-modulated Swish-gated linear unit (SwiGLU) and normalized linear head produce $\widehat A_t$.

\subsection{State-Consistent Spectral-Time Refinement}\label{sec:str}
Direct prediction supplies a nominal action chunk, but contact and state deviations can make its starting reference inconsistent with the reached robot state. STR addresses this interface through memory confidence gating, boundary reconnection, state-informed fusion, and spectral residual attenuation.

\subsubsection{Action Memory Confidence Gating}
At regeneration, let $A=\widehat A_t[s_e{:}H]$, $\mathbf a_0=A[1]$, $\mathbf d=A[H_s]-\mathbf a_0$, and $\mathbf q=\mathbf q_t$. Memory retains the measured state $\mathbf q^-$, refined terminal reference $\mathbf b$, terminal increment $\mathbf v^-$, goal displacement $\mathbf d^-$, and gates; superscripts $-$ denote stored values. With $\nu(\mathbf x)=\|\mathbf x\|_2/\sqrt{d_a}$, the discrepancies $e_m=\nu(\mathbf q-\mathbf b)$ and $e_p=\nu(\mathbf q-\mathbf a_0)$ define
\begin{equation}\label{eq:memory-gates}
\begin{aligned}
\gamma&=1-S((e_m-\delta_m)/r_m),\\
\mu&=S((e_p-\delta_p)/r_p),
\end{aligned}
\end{equation}
where $S(x)=3\bar x^2-2\bar x^3$, $\bar x=\operatorname{clip}(x,0,1)$, $(\delta_m,r_m)=(0.035,0.12)$, and $(\delta_p,r_p)=(0.025,0.12)$. Low $\gamma$ reduces reliance on memory, while $\mu$ measures mismatch with the current prediction. These are heuristic scores in separately normalized feature spaces, not physical tracking errors.

\subsubsection{Minimum Jerk Boundary Reconnection}
STR first reconciles the stored terminal reference with the nominal start through a deterministic state-reconciliation map:
\begin{equation}\label{eq:str-boundary-map}
\Phi=\mathcal R_{\mathrm{state}}(A,\mathbf q,\mathcal M_{t-1}).
\end{equation}
It uses memory confidence, state--prediction mismatch, measured progress, and plan-reversal evidence to balance stored reference and nominal intent. Unlearned gates track disturbance holdoff $h$, goal commitment $g$, turnaround $\vartheta$, unlocking $u$, and completion release $c$ at regeneration. The resulting memory gain is $\gamma'=\gamma(1-0.16c)$.

The corrected start is reconnected over $H_b=\min(10,H_s)$ rows using
\begin{equation}\label{eq:str-reconnection-window}
\begin{aligned}
m(v)&=10v^3-15v^4+6v^5,\\
w_n&=1-m\!\left(\min\{(n-1)/(H_b-1),1\}\right).
\end{aligned}
\end{equation}
The polynomial has vanishing first and second derivatives at its endpoints. The first line of Eq.~(\ref{eq:state-anchor}) reconnects the corrected start to the nominal segment; the correction tapers to zero without altering later rows.

\subsubsection{State-Informed Boundary Fusion}
Reconnection preserves nominal intent, whereas fusion blends its starting reference with normalized proprioception.
For valid memory, let $e_s=\nu(\mathbf q-\Phi)$ and let $r_j$ denote jointwise reversal release. The fusion weights are
\begin{equation}\label{eq:str-anchoring-weights}
\begin{aligned}
\kappa&=\kappa_{\max}S((e_s-\delta_a)/r_a),\\
\chi&=\prod_{i=1}^{4}(1-\alpha_i x_i),
\quad \mathbf x=(c,\vartheta,u,g),\\
c^{\mathrm{anc}}&=\max\{\kappa\mu\chi,\alpha_hh\},\\
\lambda_j&=c^{\mathrm{anc}}(1-\alpha_rr_j),
\end{aligned}
\end{equation}
where $\kappa_{\max}=0.46$ and $(\delta_a,r_a)=(0.015,0.11)$. Gate release uses $\boldsymbol\alpha=(0.75,0.65,0.55,0.60)$, with holdoff floor $\alpha_h=0.88$ and reversal coefficient $\alpha_r=0.80$. These releases reduce fusion during planned transitions. STR reconnects and fuses the start as
\begin{equation}\label{eq:state-anchor}
\begin{aligned}
A^{\mathrm{intent}}[n]&=A[n]+w_n(\Phi-\mathbf a_0),\\
\mathbf a^{\mathrm{anc}}&=(\mathbf1-\boldsymbol\lambda)\odot\Phi
+\boldsymbol\lambda\odot\mathbf q,\\
A^{\mathrm{pend}}[n]&=A^{\mathrm{intent}}[n]
+w_n(\mathbf a^{\mathrm{anc}}-\Phi).
\end{aligned}
\end{equation}
Absent memory or $\|\mathbf q-\mathbf b\|_\infty>1.5$ clears the gates and sets $\Phi=\mathbf a_0$, $\boldsymbol\lambda=\mathbf0$; spectral refinement still follows.

\subsubsection{Spectral Residual Attenuation}
STR separates the endpoint trend $B[n]=(1-v_n)A^{\mathrm{pend}}[1]+v_nA^{\mathrm{pend}}[H_s]$, where $v_n=(n-1)/(H_s-1)$. An orthonormal temporal type-II discrete cosine transform (DCT-II) gives $F=\operatorname{DCT}(A^{\mathrm{pend}}-B)$. It retains the lowest coefficients and processes higher residual frequencies as
\begin{equation}\label{eq:spectral-refinement}
F^{\mathrm{STR}}[k,:]=
\begin{cases}
F[k,:],&0\leq k<k_p,\\
\mathcal C_{\beta_{\mathrm{sp}}}(\rho_kF[k,:]),&k_p\leq k<H_s,
\end{cases}
\end{equation}
where $\rho_k=\exp[-\max(k-k_c,0)^2/(2\sigma_f^2)]$, $k_p=2$, $k_c=4$, $\sigma_f=\max(k_c/2,1)$, and $\beta_{\mathrm{sp}}=1.5$. The limiter $\mathcal C_\beta(x)$ equals $x$ for $|x|\leq\beta$ and $\beta\tanh(x/\beta)$ otherwise. This acts on residual variation rather than the endpoint trend. After $Y=B+\operatorname{IDCT}(F^{\mathrm{STR}})$, endpoint compensation restores both references:
\begin{equation}\label{eq:endpoint-compensation}
\begin{aligned}
A^{\mathrm{STR}}[n]={}&Y[n]+[1-m(v_n)](A^{\mathrm{pend}}[1]-Y[1])\\
&+m(v_n)(A^{\mathrm{pend}}[H_s]-Y[H_s]).
\end{aligned}
\end{equation}
STR queues the first $H_e$ targets and retains the intent and refined chunks and gating states for subsequent regeneration. Memory stores the measured state and the full refined chunk's terminal target. It also records the mean command increment over the final two sample intervals and the displacement from the first to the final target. The full terminal reference is retained even when $H_e<H_s$. Algorithm~\ref{alg:str} summarizes the procedure.

\begin{algorithm}[!t]
\footnotesize
\caption{DMAT Training and Action Chunk Execution}\label{alg:dmat}
\begin{algorithmic}[1]
\REQUIRE $\mathcal D,\sigma,\theta,K,H,H_e$
\ENSURE $\theta$ and joint-position commands executed in chunks
\STATE \textbf{Training:}
\REPEAT
\STATE Sample and normalize $(o_t,A_t)\sim\mathcal D$; sample $\tau\sim\mathcal U(0,1)$ and $\operatorname{vec}(E)\sim\mathcal N(\mathbf0,\sigma^2\mathbf I_{Hd_a})$
\STATE $Z_\tau\leftarrow(1-\tau)E+\tau A_t$, $\widehat A_t\leftarrow f_\theta(Z_\tau,\tau,o_t)$
\STATE $\theta\leftarrow\operatorname{AdamW}(\theta,\nabla_\theta\mathcal L_{\mathrm{DMAT}})$
\UNTIL{the configured training-update budget is reached}
\STATE \textbf{Deployment:} $s_e\leftarrow K$; initialize observation queue $\mathcal Q_o$, action queue $\mathcal Q_A$, and memory $\mathcal M_0$
\FOR{each control cycle $t=1,2,\ldots$}
\STATE $\mathcal M_t\leftarrow\mathcal M_{t-1}$
\STATE $\mathcal Q_o\leftarrow\operatorname{Update}(\mathcal Q_o,\operatorname{Normalize}(P_t,\mathbf q_t))$
\IF{$\mathcal Q_A=\varnothing$}
\STATE $o_t\leftarrow\operatorname{Stack}(\mathcal Q_o)$, $\operatorname{vec}(Z_{\mathrm{init}})\sim\mathcal N(\mathbf0,\sigma^2\mathbf I_{Hd_a})$
\STATE $\widehat A_t\leftarrow f_\theta(Z_{\mathrm{init}},0,o_t)$
\STATE $(\widetilde A_t,\mathcal M_t)\leftarrow\operatorname{STR}(\widehat A_t,\mathbf q_t,\mathcal M_{t-1},s_e)$
\STATE $\mathcal Q_A\leftarrow\operatorname{Enqueue}(\operatorname{Unnormalize}(\widetilde A_t[s_e{:}s_e+H_e-1]))$
\ENDIF
\STATE $\mathbf a_t\leftarrow\operatorname{Pop}(\mathcal Q_A)$; $\operatorname{Execute}(\mathbf a_t)$
\ENDFOR
\end{algorithmic}
\end{algorithm}

\begin{algorithm}[!t]
\footnotesize
\caption{State-Consistent Spectral-Time Refinement}\label{alg:str}
\begin{algorithmic}[1]
\REQUIRE $\widehat A_t$, $\mathbf q_t$, $\mathcal M_{t-1}$, $s_e$
\ENSURE $\widetilde A_t$, $\mathcal M_t$
\STATE $A\leftarrow\widehat A_t[s_e{:}H]$; initialize $\widetilde A_t\leftarrow\widehat A_t$
\IF{memory is absent or $\|\mathbf q_t-\mathbf b\|_\infty>1.5$}
\STATE Clear memory and gates; $\Phi\leftarrow A[1]$; $\boldsymbol\lambda\leftarrow\mathbf0$
\ELSE
\STATE Evaluate consistency, measured progress, and reversal evidence
\STATE Update holdoff, goal commitment, turnaround, unlocking, and completion sequentially
\STATE Compute the corrected start by Eq.~(\ref{eq:str-boundary-map}) and the anchoring weights
\ENDIF
\STATE Reconnect and anchor the segment by Eq.~(\ref{eq:state-anchor}); retain the intent segment reconnected to $\Phi$
\STATE Refine the spectral residual and restore endpoints by Eqs.~(\ref{eq:spectral-refinement}) and~(\ref{eq:endpoint-compensation})
\STATE $\widetilde A_t[s_e{:}H]\leftarrow A^{\mathrm{STR}}$; with valid memory, blend unqueued history rows as $\gamma'\mathbf b_{\mathrm{old}}+(1-\gamma')\widehat{\mathbf a}_{t,n}$
\STATE Store both chunks, measured state, terminal reference, motion increment, goal displacement, and gates in $\mathcal M_t$
\STATE \textbf{return} $\widetilde A_t,\mathcal M_t$
\end{algorithmic}
\end{algorithm}

\section{Experiments and Results}\label{sec:experiments}

\subsection{Implementation Details}\label{sec:implementation}
\subsubsection{Robot Platform and Tasks}
Experiments use a Fourier GR2 upper body bimanual robot with 29 proprioceptive positions: 14 arm, 12 hand, two head, and one fixed waist joint. A head-mounted RealSense L515 supplies pointclouds cropped to the workspace with the table removed. The first $H_e$ refined targets enter a command queue for the position servo; an empty queue triggers updated sensing and prediction, placing STR at the learned-policy--servo interface.

The tasks probe complementary demands on this perception-to-execution chain.

\textbf{Pick Place (basic rigid-object manipulation)} requires grasping a randomly initialized object and placing it in a designated container, testing localization, grasp closure, transport, and release.

\textbf{Vase Wipe (complex curved-surface contact)} requires retaining a wiping tool, maintaining contact with a nonplanar vase, and removing predefined ink marks. Varying surface normals, curvature, local occlusion, and accumulated error require coordinated motion without detachment or jamming. The task therefore tests sustained-contact execution without force or tactile sensing.

\textbf{Pour Beans (long-horizon bimanual coordination and dynamic-medium manipulation)} requires one arm to pour while the other stabilizes the receiving container. Relative-pose error causes spilling, making the task a test of long-horizon coordination with a dynamic medium \cite{schenck2017learning}.

\textbf{Data collection and configuration.} Expert demonstrations were collected with Apple Vision Pro teleoperation. Each task contains 50 successful trajectories shared by all methods. For evaluation, the $15\,\mathrm{cm}\times15\,\mathrm{cm}$ right-hand workspace is divided into a $3\times3$ grid for object initialization; the left-hand object remains fixed in bimanual tasks.

\subsubsection{Baselines and Metrics}
Pick Place and Vase Wipe use $(K,H,H_e)=(2,16,15)$, where $K$, $H$, and $H_e$ denote observation history length, predicted horizon, and queued executable targets; Pour Beans uses $(2,64,32)$. DMAT uses one network evaluation and unit noise scale; AdamW uses a learning rate of $10^{-4}$ and batch size 64. Latency was measured on a single NVIDIA GeForce RTX 4090 GPU at batch size one.

We compare DMAT with 3D diffusion and flow matching policies under shared data and physical protocols. IDP3 iteratively denoises joint-space chunks from pointclouds and proprioception \cite{ze2025generalizable}. DiTFlow3D combines the corresponding pointcloud encoder with a Diffusion Transformer (DiT) velocity predictor and Euler integration \cite{lipman2022flow}, \cite{dasari2024dit}; both baselines use ten steps.

We use task success rate, action chunk generation latency, trajectory smoothness, action command fluctuation, and action command redundancy. The last three are evaluated on Vase Wipe. SPARC values closer to zero indicate smoother motion, while smaller fluctuation and redundancy indicate more regular commands.
\begin{enumerate}
\item \textbf{Task success rate.} Pick Place requires stable placement, Vase Wipe requires removal of at least 90\% of visible ink, and Pour Beans requires continuous pouring without obvious large-area spilling. The same predefined criteria are applied to every method and retained in OOD tests; intervention trials additionally require task completion despite the prescribed 50 N pulls. The aggregate reported in the abstract combines 108 standard trials and 459 OOD trials per method; intervention trials are reported separately.

\item \textbf{Action chunk generation latency.} Generation time is recorded whenever a chunk is updated during 1000 control steps; valid calls are averaged per task, and the three task means are weighted equally.

\item \textbf{Trajectory smoothness.} Spectral arc length (SPARC) is computed from feedback joint-speed profiles for the 17 waist, head, and arm joints after discarding the first ten logged samples. It quantifies motion smoothness through the arc length of the normalized Fourier magnitude spectrum, with values closer to zero indicating smoother motion \cite{balasubramanian2011robust}. For a nonzero profile $s_j(\xi)=|v_j(\xi)|$,
\begin{equation}\label{eq:sparc}
\mathrm{SPARC}_j=-\int_0^{\omega_{c,j}}\sqrt{\frac{1}{\omega_{c,j}^2}+\left(\frac{d\hat S_j(\omega)}{d\omega}\right)^2}\,d\omega,
\end{equation}
where $\omega$ denotes frequency in Hz, $S_j$ is the Fourier magnitude spectrum of $s_j$, $\hat S_j(\omega)=S_j(\omega)/\max_{\omega'}S_j(\omega')$ and $\omega_{c,j}=\min(10\,\mathrm{Hz},\omega_{\alpha,j})$, with $\omega_{\alpha,j}=\max\{\omega:\hat S_j(\omega)\geq0.05\}$. Profiles with fewer than two samples, peak magnitude below $10^{-6}$, or a nonpositive cutoff are undefined and excluded; the remaining joint scores are averaged within each rollout. All signal-valid started trials, including failed trials, remain in the summary.

\item \textbf{Action command fluctuation.} This metric quantifies local variation in successive increments of the target joint commands within each action chunk. For Vase Wipe, let $a_{r,k,n,j}$ be the target in degrees for arm joint $j$ at row $n$ of complete executed segment $k$ in rollout $r$. With $K_r$ complete segments and $H_e=15$, the trajectory-level mean absolute second difference is
\begin{equation}
\label{eq:action-command-fluctuation}
\begin{aligned}
C_r&=\frac{1}{14K_r(H_e-2)}\sum_{k=1}^{K_r}\sum_{n=2}^{H_e-1}\sum_{j=1}^{14}\\
&\quad \left|a_{r,k,n+1,j}-2a_{r,k,n,j}+a_{r,k,n-1,j}\right|.
\end{aligned}
\end{equation}
Lower $C_r$ values indicate more regular command increments; $C_r=0$ corresponds to zero second difference within the evaluated segments.

\item \textbf{Action command redundancy.} This metric measures back-and-forth command travel beyond a segment's net displacement. For each segment--joint pair,
\begin{equation}
\label{eq:action-command-path}
\begin{aligned}
L_{r,k,j}
&=\sum_{n=1}^{H_e-1}
\left|a_{r,k,n+1,j}-a_{r,k,n,j}\right|,\\
D_{r,k,j}
&=\left|a_{r,k,H_e,j}-a_{r,k,1,j}\right|.
\end{aligned}
\end{equation}
For pairs with total variation $L_{r,k,j}\geq\eta$, the normalized excess path is
\begin{equation}
\label{eq:action-command-redundancy}
\begin{aligned}
E_r&=\frac{1}{|\mathcal V_r|}\sum_{(k,j)\in\mathcal V_r}\frac{L_{r,k,j}-D_{r,k,j}}{L_{r,k,j}},\\
\mathcal V_r&=\{(k,j):L_{r,k,j}\geq\eta\},
\end{aligned}
\end{equation}
where $\eta=0.5^{\circ}$. Lower $E_r$ values indicate less redundant command travel; $E_r=0$ means that total variation equals net displacement, with no excess back-and-forth path. Incomplete segments, rollouts with $K_r=0$, and redundancy cases with $\mathcal V_r=\varnothing$ are excluded from the corresponding command metric. Both command summaries retain all started trials that satisfy these signal-level rules.
\end{enumerate}

\subsection{Standard Task Performance}\label{sec:standard-performance}
\begin{figure}[!t]
\centering
\includegraphics[width=\columnwidth]{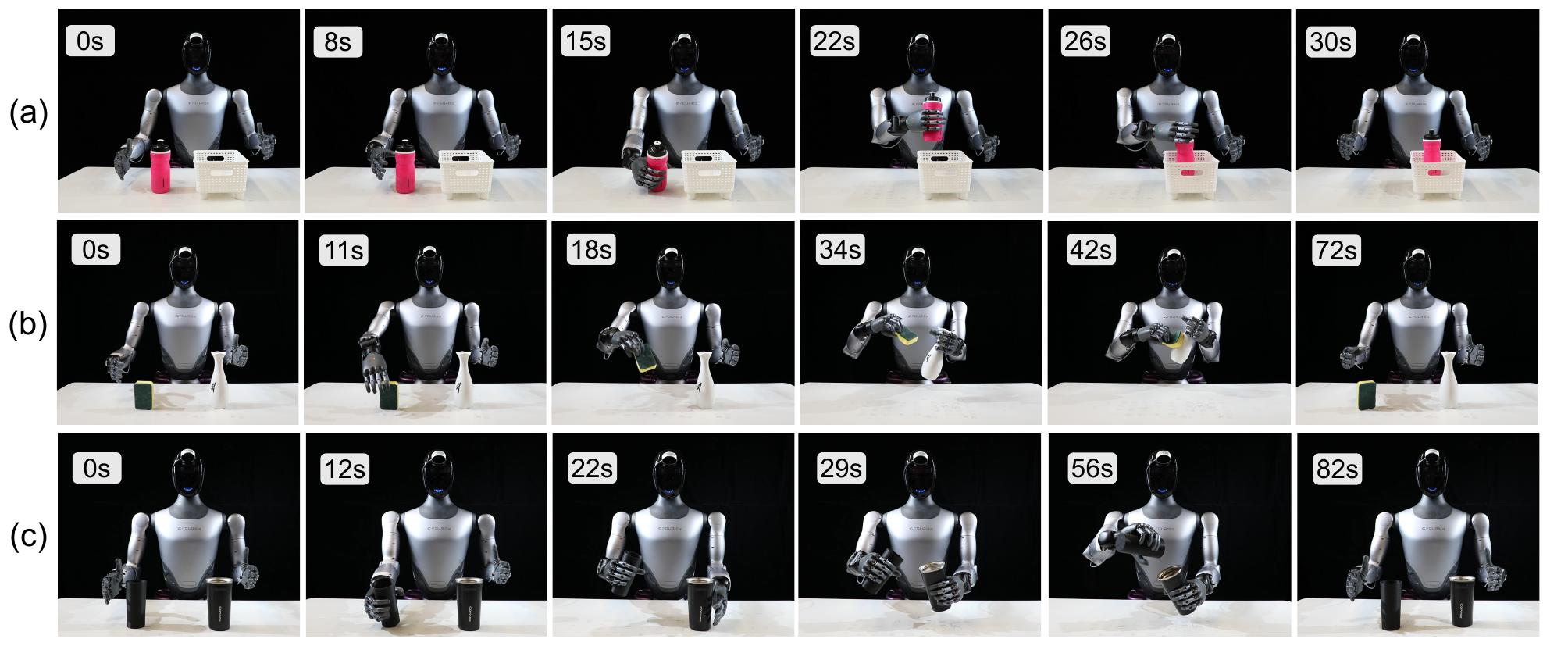}
\caption{Representative successful rollouts under standard conditions: (a) Pick Place, (b) Vase Wipe, and (c) Pour Beans. Rows show time-ordered frames with elapsed time.}
\label{fig:standard-sequences}
\end{figure}

Standard trials first test grasp transfer, sustained wiping contact, and coordinated pouring, with 36 trials per task and method. Representative successful sequences show how these task-level outcomes are realized (Fig.~\ref{fig:standard-sequences}).

\begin{figure}[!t]
\centering
\includegraphics[width=\columnwidth]{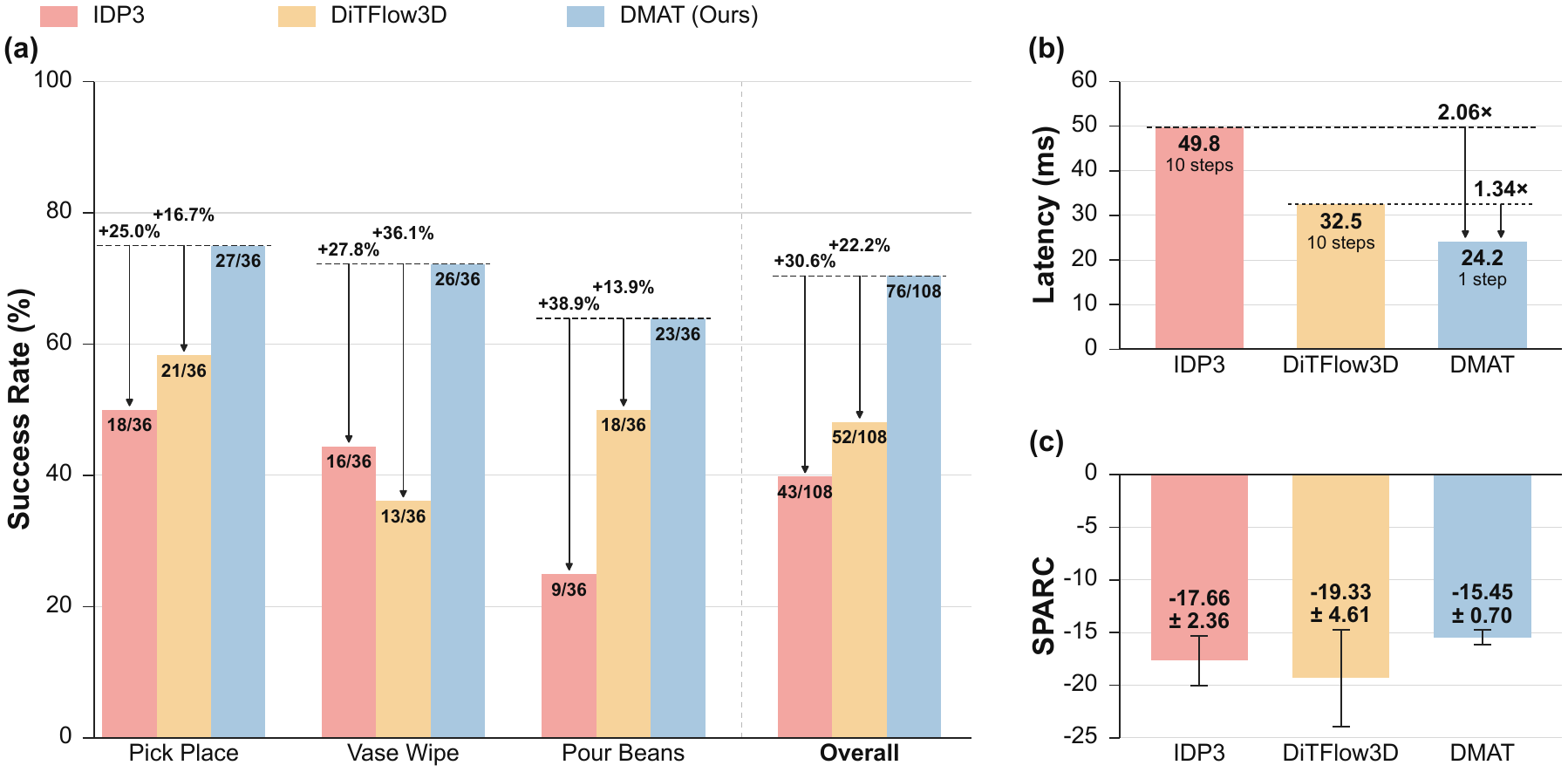}
\caption{Standard condition comparison. (a) Task success on Pick Place, Vase Wipe, Pour Beans, and the aggregate, with 36 trials per task; arrow labels indicate the absolute differences in success rate between DMAT (w/ STR) and the comparison methods. (b) Equal-weight average action chunk generation latency; labels indicate latency and inference steps, and arrows indicate DMAT speedups. (c) Vase Wipe joint-speed SPARC (mean $\pm$ standard deviation).}
\label{fig:standard-performance}
\end{figure}

The quantitative standard condition comparison then shows that DMAT achieved the highest success on each task and 70.4\% aggregate success (76/108), corresponding to relative improvements of 76.7\% over IDP3 and 46.2\% over DiTFlow3D (Fig.~\ref{fig:standard-performance}(a)).

The same comparison quantifies generation cost and measured motion. DMAT generated each chunk in 24.2 ms, giving $2.06\times$ and $1.34\times$ speedups over IDP3 and DiTFlow3D, while its Vase Wipe SPARC was $-15.45\pm0.70$, compared with $-17.66\pm2.36$ and $-19.33\pm4.61$ (Fig.~\ref{fig:standard-performance}(b) and (c)).

Within each chunk, command metrics complement the feedback-motion measurements (Table~\ref{tab:standard-command-quality}). DMAT produced the lowest Vase Wipe fluctuation and redundancy, decreasing them by 77.0\% and 50.3\% relative to IDP3 and by 85.6\% and 64.1\% relative to DiTFlow3D.

\begin{table}[!t]
\caption{Action Command Quality Under Standard Conditions}
\label{tab:standard-command-quality}
\centering
\footnotesize
\begin{tabular*}{0.94\columnwidth}{@{\extracolsep{\fill}}lcc@{}}
\toprule
\textbf{Method} & \textbf{Fluctuation (deg) $\downarrow$} & \textbf{Redundancy $\downarrow$}\\
\midrule
IDP3 & $0.1662\pm0.0119$ & $0.4134\pm0.0095$\\
DiTFlow3D & $0.2662\pm0.0167$ & $0.5716\pm0.0366$\\
DMAT (Ours) & {\boldmath$0.0383\pm0.0019$} & {\boldmath$0.2053\pm0.0147$}\\
\bottomrule
\end{tabular*}
\end{table}

\subsection{Out-of-Distribution Generalization}\label{sec:ood}

\begin{figure}[!t]
\centering
\includegraphics[width=\columnwidth]{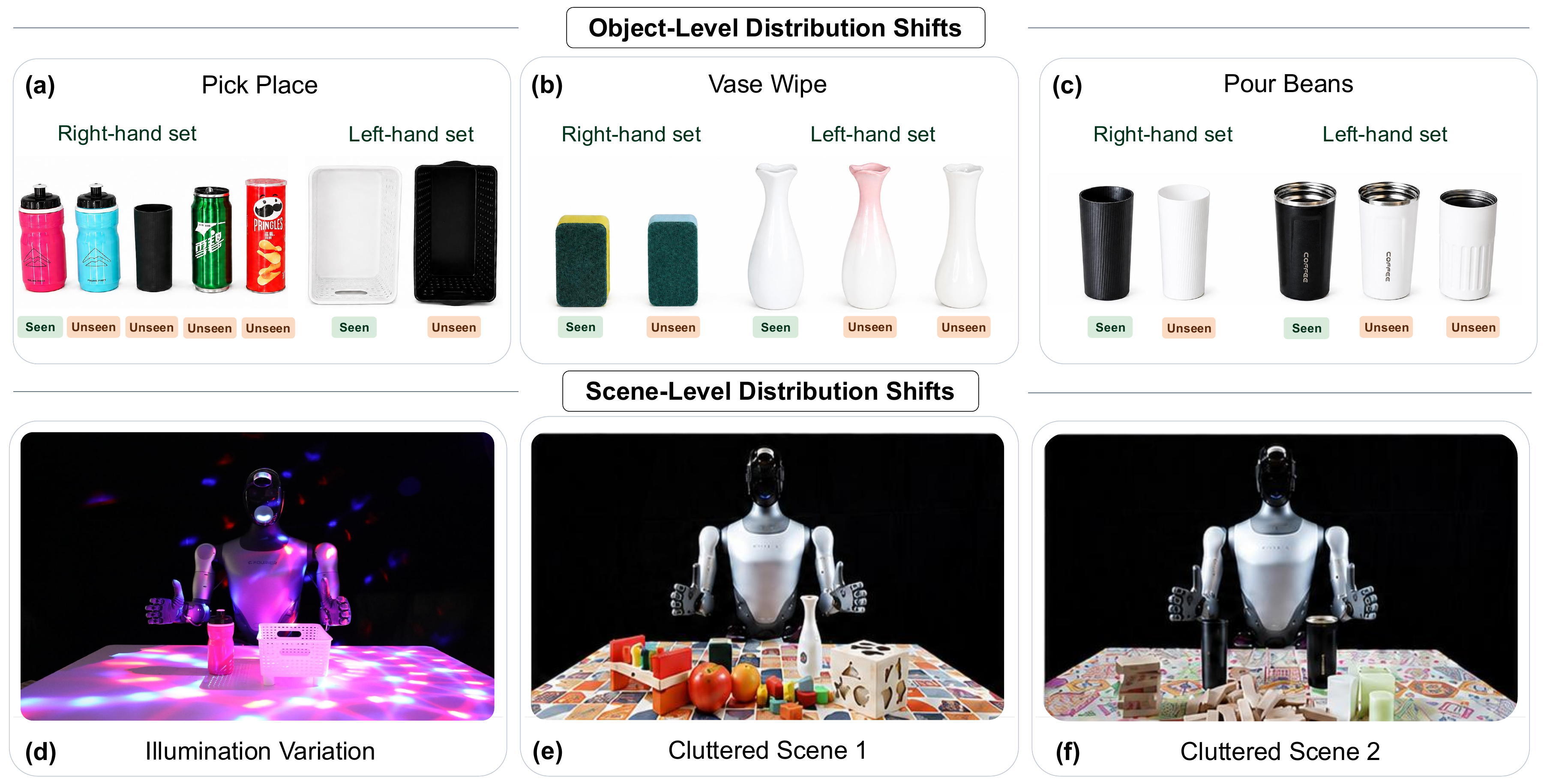}
\caption{OOD evaluation under object-level and scene-level shifts. Panels (a), (b), and (c) show seen and unseen objects for Pick Place, Vase Wipe, and Pour Beans, respectively; panels (d), (e), and (f) show illumination variation and two cluttered scenes, with no retraining or fine-tuning.}
\label{fig:ood-settings}
\end{figure}

The OOD study tests each fixed policy under specified visual and geometric shifts (Fig.~\ref{fig:ood-settings}). Illumination variation, two cluttered scenes, and unseen objects alter appearance or geometry \cite{ze20243d}, \cite{ze2025generalizable} while preserving the robot, inputs, actions, task goals, initialization, and success criteria. The scope is limited to the enumerated objects and scenes. Each method performs 27 trials per task under illumination and 54 across the cluttered scenes, while unseen-object tests contain 108 Pick Place trials and 54 trials for each bimanual task, yielding 189, 135, and 135 task-specific trials per policy.

\begin{figure}[!t]
\centering
\includegraphics[width=\columnwidth]{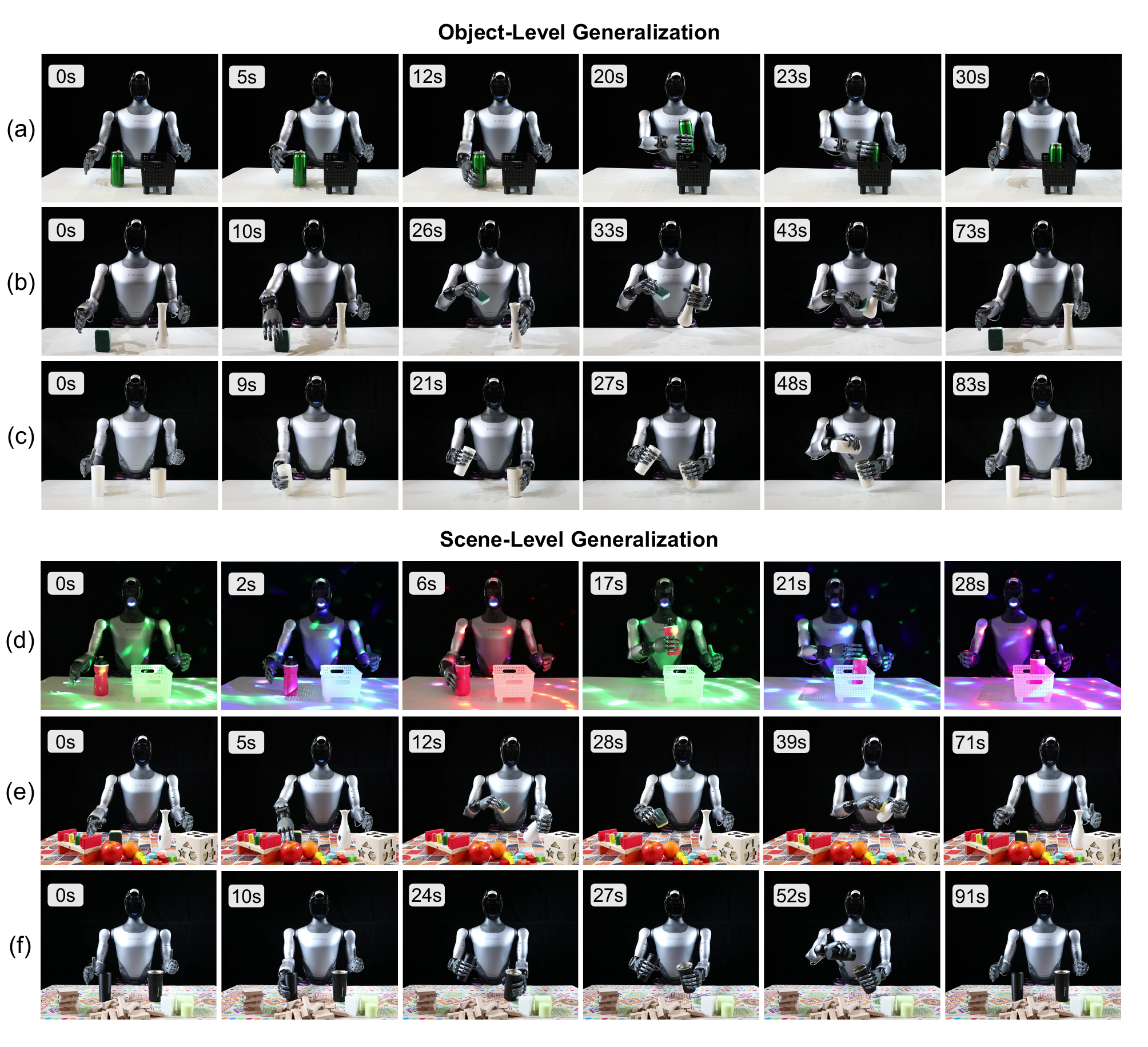}
\caption{Representative successful DMAT rollouts under OOD conditions without retraining or fine-tuning. Rows (a)--(c) show unseen-object generalization for Pick Place, Vase Wipe, and Pour Beans, respectively, while rows (d)--(f) show scene-level shifts for the same tasks. Each row presents six time-ordered frames from a single successful rollout.}
\label{fig:ood-sequences}
\end{figure}

Representative sequences show DMAT execution under unseen-object and scene-level shifts without retraining or fine-tuning (Fig.~\ref{fig:ood-sequences}).

\begin{figure*}[!t]
\centering
\includegraphics[width=0.96\textwidth]{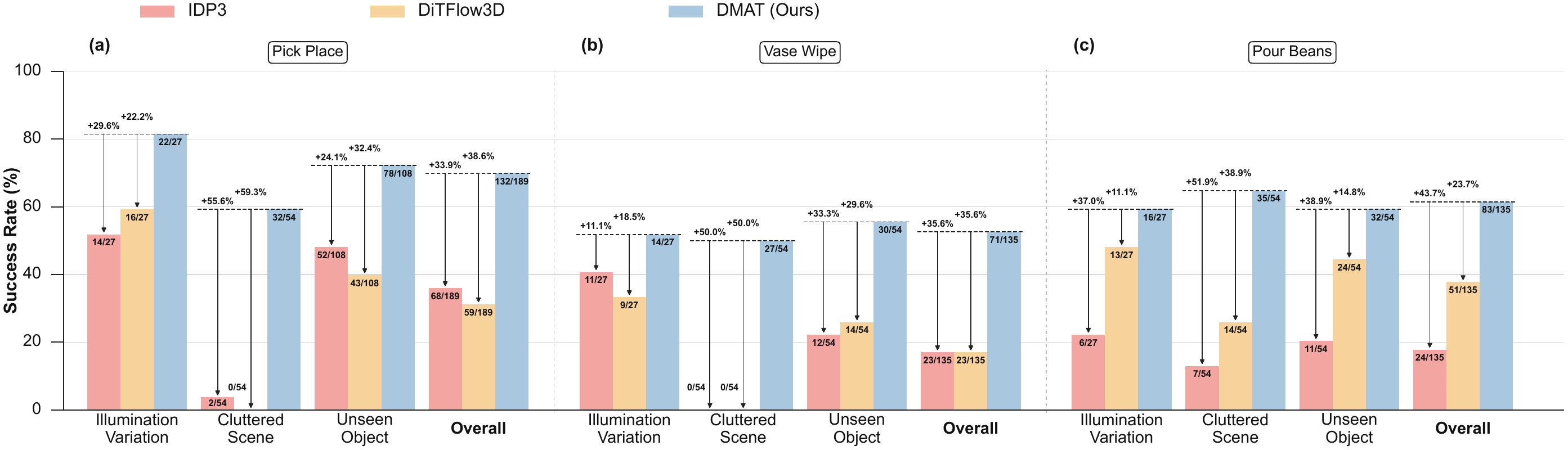}
\caption{Task success under visual and geometric OOD shifts. Panels (a)--(c) show illumination, clutter, unseen object, and aggregate results for Pick Place, Vase Wipe, and Pour Beans, respectively. No policy is retrained or fine-tuned for these OOD evaluations. Arrow labels indicate the absolute differences in success rate between DMAT (w/ STR) and the comparison methods.}
\label{fig:ood-performance}
\end{figure*}

Across the specified shifts, DMAT achieved the highest aggregate success for each task and 62.3\% overall (286/459), corresponding to relative improvements of 148.7\% over IDP3 and 115.0\% over DiTFlow3D (Fig.~\ref{fig:ood-performance}). Clutter produced the clearest separation among the evaluated policies: DMAT completed 32/54, 27/54, and 35/54 Pick Place, Vase Wipe, and Pour Beans trials, respectively, whereas IDP3 completed 2/54, 0/54, and 7/54 and DiTFlow3D completed 0/54, 0/54, and 14/54.

\subsection{Robustness Under Physical Interventions}\label{sec:physical-intervention}
The intervention study tests whether STR helps a generated chunk remain executable after an induced state deviation. During Vase Wipe, an operator pulls laterally through a digital force gauge and releases at its audible 50 N threshold. Three pulls occur at preset times, or 30 within a preset interval, with nine trials per policy and condition. STR acts at the next regeneration boundary; DMAT without STR shares the same nominal predictor, and both baselines follow the same protocol.

\begin{figure*}[!t]
\centering
\includegraphics[width=0.95\textwidth]{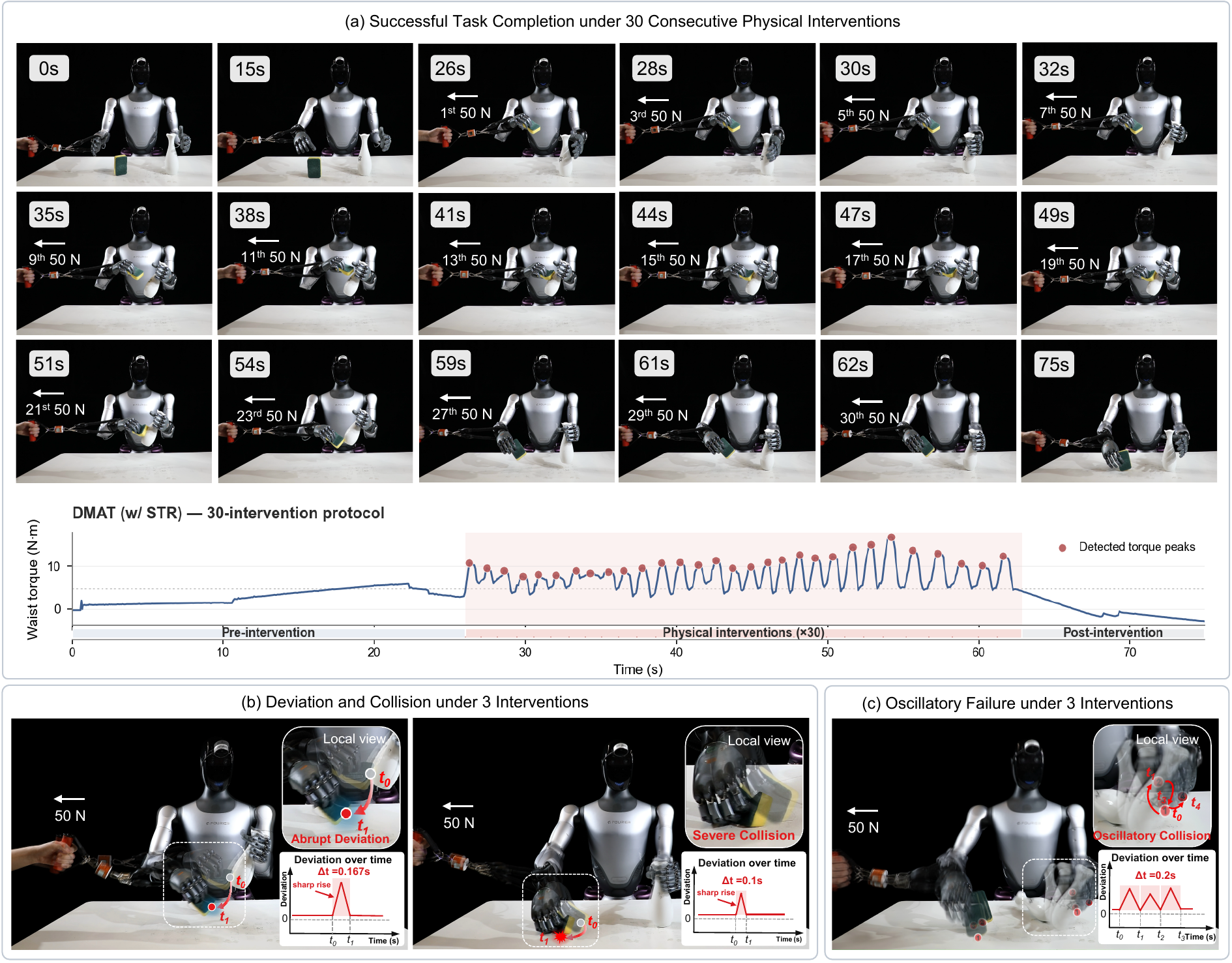}
\caption{Representative responses to 50 N lateral-pull interventions. (a) DMAT completes Vase Wipe after 30 interventions. Under the three-intervention condition, (b) IDP3 deviates abruptly and collides, whereas (c) DiTFlow3D oscillates repeatedly and fails. Motion overlays and local views highlight the responses; white arrows and labels denote pulls reaching the 50 N threshold.}
\label{fig:intervention-examples}
\end{figure*}

Representative responses provide execution context (Fig.~\ref{fig:intervention-examples}). Full DMAT resumed wiping and completed the selected 30-pull trial, whereas the selected three-pull IDP3 and DiTFlow3D trials ended in collision and oscillatory failure, respectively.

\begin{figure}[!t]
\centering
\includegraphics[width=\columnwidth]{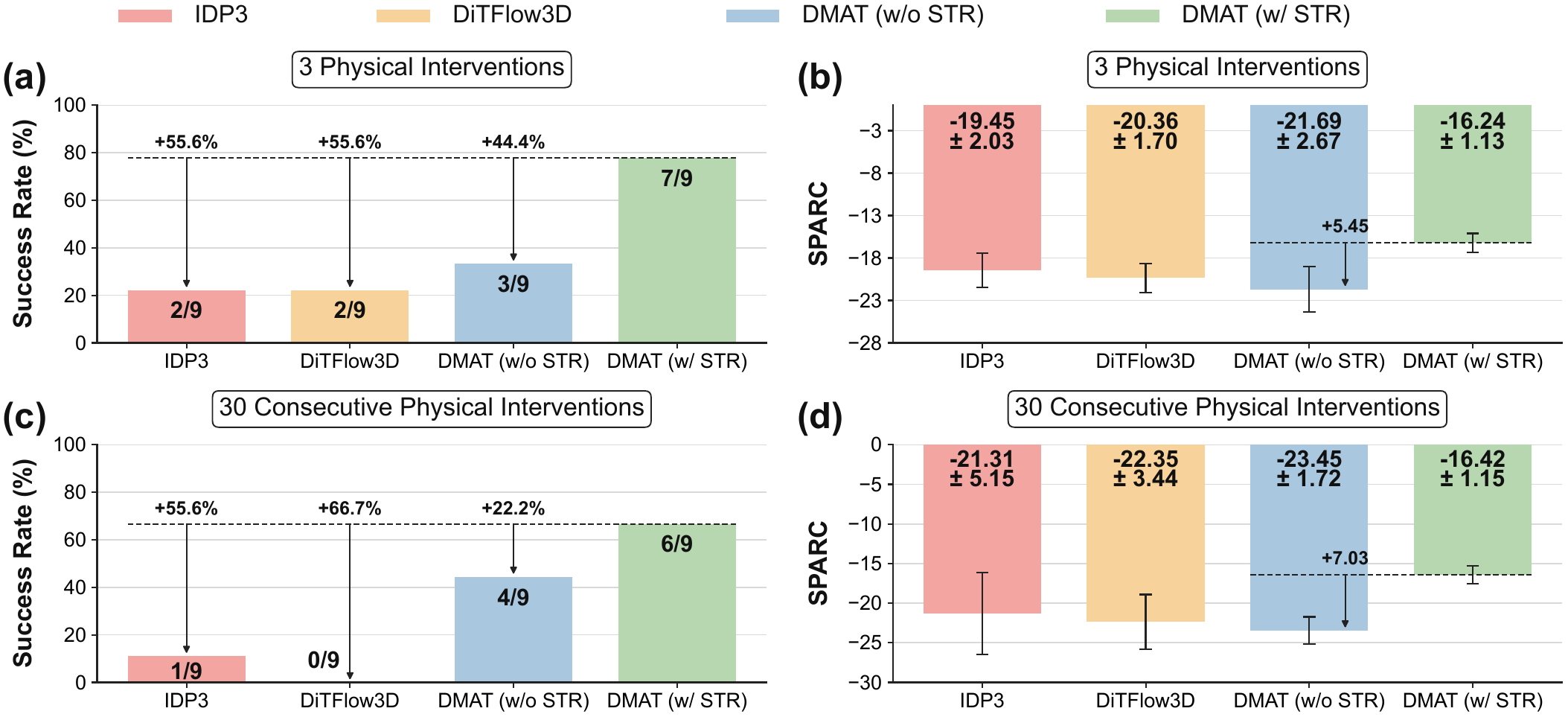}
\caption{Post-intervention task success and motion smoothness under 50 N lateral-pull interventions. (a) and (c) show Vase Wipe success under 3 and 30 interventions, respectively; arrow labels indicate the absolute differences in success rate between DMAT (w/ STR) and the comparison methods. (b) and (d) show joint-speed SPARC (mean $\pm$ standard deviation).}
\label{fig:intervention-performance}
\end{figure}

Under the prescribed intervention protocol, STR improved observed task completion and feedback-motion smoothness (Fig.~\ref{fig:intervention-performance}). On the same DMAT backbone, STR increased post-intervention completion from 3/9 to 7/9 under three interventions and from 4/9 to 6/9 under 30 interventions. IDP3 and DiTFlow3D each completed 2/9 trials under three interventions; under 30 interventions, they completed 1/9 and 0/9 trials, respectively. STR also shifted SPARC from $-21.69\pm2.67$ to $-16.24\pm1.13$ under three interventions and from $-23.45\pm1.72$ to $-16.42\pm1.15$ under 30 interventions, indicating smoother measured joint motion under this protocol.

Command results show that full DMAT produced the lowest fluctuation and redundancy under both intervention conditions (Table~\ref{tab:intervention-command-quality}). On the same DMAT backbone, STR reduced fluctuation from $0.103\pm0.005$ to $0.038\pm0.002$ degrees and redundancy from $0.301\pm0.015$ to $0.217\pm0.007$ under three interventions, corresponding to relative reductions of 63.1\% and 27.9\%, respectively. Under 30 interventions, fluctuation decreased from $0.119\pm0.012$ to $0.038\pm0.002$ degrees and redundancy from $0.324\pm0.016$ to $0.216\pm0.024$, corresponding to reductions of 68.1\% and 33.3\%.

\begin{table}[!t]
\caption{Action Command Quality Under 50 N Interventions}
\label{tab:intervention-command-quality}
\centering
\footnotesize
\begin{tabular*}{0.94\columnwidth}{@{\extracolsep{\fill}}lcc@{}}
\toprule
\textbf{Method} & \textbf{Fluctuation (deg) $\downarrow$} & \textbf{Redundancy $\downarrow$}\\
\midrule
\multicolumn{3}{c}{\textbf{Three interventions}}\\
\midrule
IDP3 & $0.176\pm0.018$ & $0.436\pm0.016$\\
DiTFlow3D & $0.320\pm0.042$ & $0.595\pm0.060$\\
DMAT (w/o STR) & $0.103\pm0.005$ & $0.301\pm0.015$\\
DMAT (w/ STR) & {\boldmath$0.038\pm0.002$} & {\boldmath$0.217\pm0.007$}\\
\midrule
\multicolumn{3}{c}{\textbf{30 interventions}}\\
\midrule
IDP3 & $0.188\pm0.020$ & $0.421\pm0.021$\\
DiTFlow3D & $0.341\pm0.060$ & $0.663\pm0.023$\\
DMAT (w/o STR) & $0.119\pm0.012$ & $0.324\pm0.016$\\
DMAT (w/ STR) & {\boldmath$0.038\pm0.002$} & {\boldmath$0.216\pm0.024$}\\
\bottomrule
\end{tabular*}
\end{table}

\subsection{Ablation Study}\label{sec:ablation}
Progressive ablation evaluates the incremental contributions of geometry and STR (Table~\ref{tab:ablation-performance}). DMAT-Direct contains the direct predictor, DMAT-Geo adds unpooled spatial tokens and cross-attention without STR, and full DMAT adds STR. These variants completed 48/108, 69/108, and 76/108 trials at 8.69, 22.95, and 24.20 ms, respectively. On Vase Wipe, STR increased completion from 21/36 to 26/36. The direct predictor had the lowest latency, while geometry and STR successively increased observed task success.

\begin{table}[!t]
\caption{Ablation of DMAT Components}\label{tab:ablation-performance}
\centering
\scriptsize
\setlength{\tabcolsep}{2pt}
\begin{tabular}{@{}lcccccc@{}}
\toprule
\textbf{Variant} 
& \textbf{\begin{tabular}{@{}c@{}}Pick\\Place\end{tabular}} 
& \textbf{\begin{tabular}{@{}c@{}}Vase\\Wipe\end{tabular}} 
& \textbf{\begin{tabular}{@{}c@{}}Pour\\Beans\end{tabular}} 
& \textbf{\begin{tabular}{@{}c@{}}Avg.\\latency\end{tabular}} 
& \textbf{\begin{tabular}{@{}c@{}}Mean SPARC on\\Vase Wipe $\uparrow$\end{tabular}} 
& \textbf{Overall}\\
\midrule
DMAT-Direct & 22/36 & 12/36 & 14/36 & \textbf{8.69 ms} & $-21.46\pm2.83$ & 48/108 (44.4\%)\\
DMAT-Geo    & 26/36 & 21/36 & 22/36 & 22.95 ms & $-20.45\pm2.63$ & 69/108 (63.9\%)\\
DMAT (Full) & \textbf{27/36} & \textbf{26/36} & \textbf{23/36} & 24.20 ms & {\boldmath$-15.45\pm0.70$} & \textbf{76/108 (70.4\%)}\\
\bottomrule
\end{tabular}
\end{table}

Motion and command metrics further distinguish geometry from state-consistent execution (Table~\ref{tab:ablation-command-quality}). Vase Wipe SPARC progressed from $-21.46\pm2.83$ to $-20.45\pm2.63$ and $-15.45\pm0.70$. Relative to DMAT-Direct, DMAT-Geo reduced fluctuation and redundancy by 11.0\% and 18.9\%. Adding STR to DMAT-Geo reduced them by 60.8\% and 29.6\%. In this progression, geometry produced the larger task-success gain, while STR produced the larger motion and command-regularity improvements.

\begin{table}[!t]
\caption{Ablation Results for Action Command Quality}
\label{tab:ablation-command-quality}
\centering
\footnotesize
\begin{tabular*}{0.94\columnwidth}{@{\extracolsep{\fill}}lcc@{}}
\toprule
\textbf{Variant} & \textbf{Fluctuation (deg) $\downarrow$} & \textbf{Redundancy $\downarrow$}\\
\midrule
DMAT-Direct & $0.109\pm0.011$ & $0.359\pm0.059$\\
DMAT-Geo & $0.097\pm0.004$ & $0.291\pm0.010$\\
DMAT (Full) & {\boldmath$0.038\pm0.002$} & {\boldmath$0.205\pm0.015$}\\
\bottomrule
\end{tabular*}
\end{table}

\subsection{Failure Analysis}\label{sec:failure-cases}
\begin{figure}[!t]
\centering
\includegraphics[width=0.8\columnwidth]{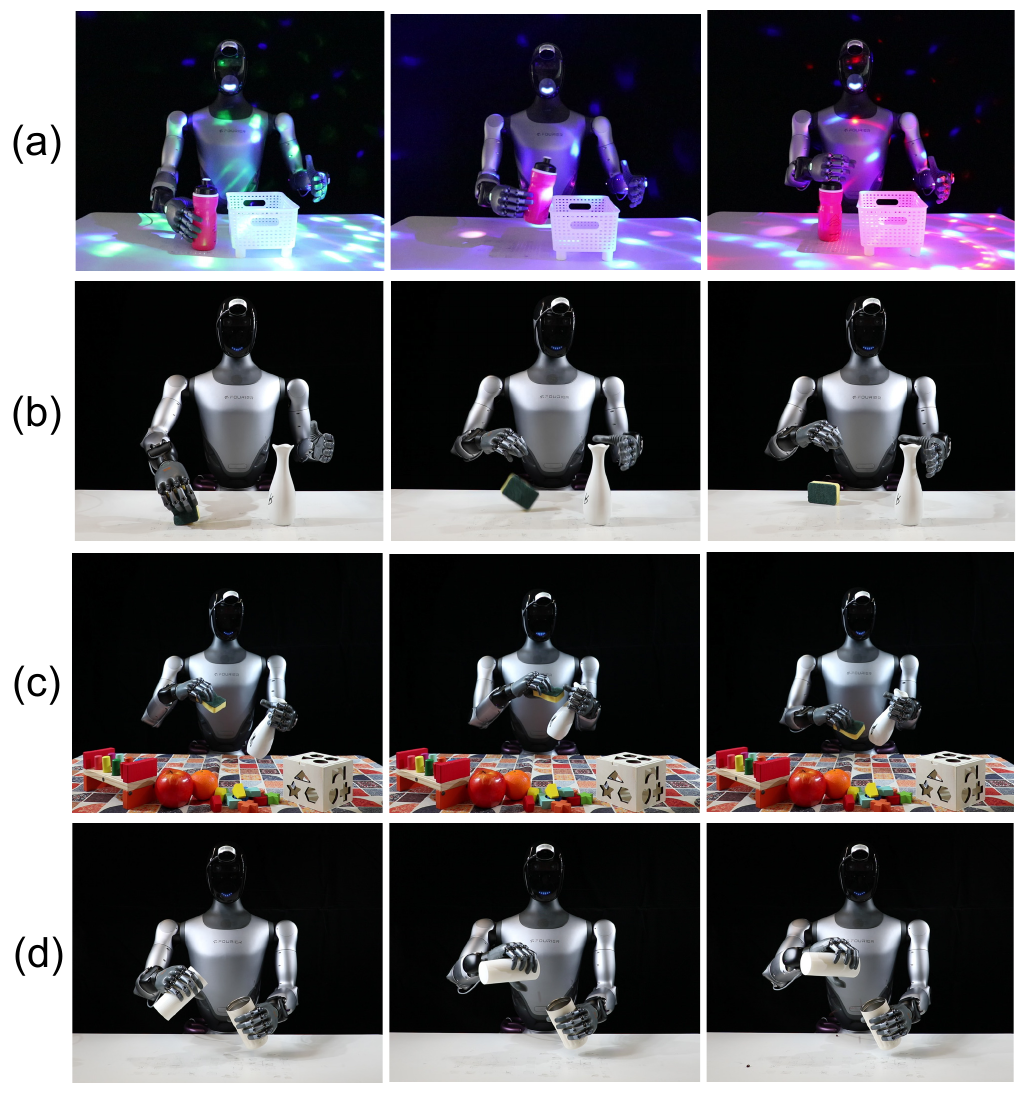}
\caption{Representative baseline failure cases under standard and OOD conditions. (a) Pick Place misses the grasp. (b) The wiping tool drops during transport. (c) Visible ink remains after wiping. (d) Beans spill during pouring.}
\label{fig:failure-cases}
\end{figure}

Representative baseline failures further clarify the challenges addressed by DMAT. Representative failures included grasp failure under partial occlusion in Pick Place, tool dropping during transport and incomplete wiping in Vase Wipe, and bean spilling during Pour Beans (Fig.~\ref{fig:failure-cases}). These failures indicate two recurring difficulties in physical manipulation: insufficient spatial grounding under occlusion or complex local geometry, and execution inconsistency caused by accumulated state deviation or oscillatory action transitions. The local geometry-anchored action space decoder addresses the former by preserving spatially resolved 3D cues for action prediction, while STR addresses the latter by reconnecting regenerated action chunks to the measured robot state and suppressing abrupt variations. Their complementary roles allow DMAT to combine efficient clean-action prediction with spatially grounded and state-consistent physical execution.

\section{Conclusion}\label{sec:conclusion}
This article presented DMAT, a direct action-manifold modeling framework for efficient and robust bimanual manipulation. Under the conditional action-manifold hypothesis, DMAT treats the clean observation-conditioned action chunk as the primary modeling target, rather than recovering it through an iterative generative path. Local geometry-anchored decoding preserves spatially resolved 3D information during action prediction, while STR reconciles regenerated actions with the measured robot state during action chunk execution. Real-robot experiments across standard, OOD, and physical-intervention settings validate the effectiveness of this formulation. The ablations further show that the performance gain does not arise from single-step inference alone: local geometric conditioning contributes the major improvement in task success, whereas STR further improves motion smoothness, command quality, and execution consistency under physical interventions. These findings support a broader modeling perspective in which efficient robot action generation is achieved by directly modeling the executable action structure and explicitly preserving the spatial and state constraints required for physical execution.

\bibliographystyle{IEEEtran}
\bibliography{DMAT_References}

@inproceedings{zhao2023learning,
  author       = {Tony Z. Zhao and
                  Vikash Kumar and
                  Sergey Levine and
                  Chelsea Finn},
  title        = {Learning Fine-Grained Bimanual Manipulation with Low-Cost Hardware},
  booktitle    = {Robotics: Science and Systems},
  year         = {2023}
}

@article{kim2024goal,
  title={Goal-conditioned dual-action imitation learning for dexterous dual-arm robot manipulation},
  author={Kim, Heecheol and Ohmura, Yoshiyuki and Kuniyoshi, Yasuo},
  journal={IEEE Transactions on Robotics},
  volume={40},
  pages={2287--2305},
  year={2024},
  publisher={IEEE}
}

@article{chi2025diffusion,
  title={Diffusion policy: Visuomotor policy learning via action diffusion},
  author={Chi, Cheng and Xu, Zhenjia and Feng, Siyuan and Cousineau, Eric and Du, Yilun and Burchfiel, Benjamin and Tedrake, Russ and Song, Shuran},
  journal={The International Journal of Robotics Research},
  volume={44},
  number={10-11},
  pages={1684--1704},
  year={2025},
  publisher={Sage Publications Sage UK: London, England}
}

@inproceedings{lipman2022flow,
  author       = {Yaron Lipman and
                  Ricky T. Q. Chen and
                  Heli Ben{-}Hamu and
                  Maximilian Nickel and
                  Matthew Le},
  title        = {Flow Matching for Generative Modeling},
  booktitle    = {{ICLR}},
  publisher    = {OpenReview.net},
  year         = {2023}
}

@article{ding2025fast,
  title={Fast and robust visuomotor riemannian flow matching policy},
  author={Ding, Haoran and Jaquier, No{\'e}mie and Peters, Jan and Rozo, Leonel},
  journal={IEEE Transactions on robotics},
  year={2025},
  publisher={IEEE}
}

@inproceedings{prasad2024consistency,
  author       = {Aaditya Prasad and
                  Kevin Lin and
                  Jimmy Wu and
                  Linqi Zhou and
                  Jeannette Bohg},
  title        = {Consistency Policy: Accelerated Visuomotor Policies via Consistency
                  Distillation},
  booktitle    = {Robotics: Science and Systems},
  year         = {2024}
}

@inproceedings{zhang2025flowpolicy,
  title={Flowpolicy: Enabling fast and robust 3d flow-based policy via consistency flow matching for robot manipulation},
  author={Zhang, Qinglun and Liu, Zhen and Fan, Haoqiang and Liu, Guanghui and Zeng, Bing and Liu, Shuaicheng},
  booktitle={Proceedings of the AAAI Conference on Artificial Intelligence},
  volume={39},
  number={14},
  pages={14754--14762},
  year={2025}
}

@inproceedings{black2025real,
  author       = {Kevin Black and
                  Manuel Y. Galliker and
                  Sergey Levine},
  title        = {Real-Time Execution of Action Chunking Flow Policies},
  booktitle    = {NeurIPS},
  year         = {2025}
}

@inproceedings{ze20243d,
  author       = {Yanjie Ze and
                  Gu Zhang and
                  Kangning Zhang and
                  Chenyuan Hu and
                  Muhan Wang and
                  Huazhe Xu},
  title        = {3D Diffusion Policy: Generalizable Visuomotor Policy Learning via
                  Simple 3D Representations},
  booktitle    = {Robotics: Science and Systems},
  year         = {2024}
}

@inproceedings{ze2025generalizable,
  title={Generalizable humanoid manipulation with 3d diffusion policies},
  author={Ze, Yanjie and Chen, Zixuan and Wang, Wenhao and Chen, Tianyi and He, Xialin and Yuan, Ying and Peng, Xue Bin and Wu, Jiajun},
  booktitle={2025 IEEE/RSJ International Conference on Intelligent Robots and Systems (IROS)},
  pages={2873--2880},
  year={2025},
  organization={IEEE}
}

@article{wang2026arvip,
  title={ARViP: Adversarial Regularization in Visuomotor Policies for Robotic VLA Purpose},
  author={Wang, Haoji and Wu, Bo and Zheng, Suibing},
  journal={IEEE Transactions on Industrial Informatics},
  year={2026},
  publisher={IEEE}
}

@article{zhao2026staged,
  title={Staged Modulation Diffusion Policy With Complementary Visual Fusion for Robotic Workpiece Assembly},
  author={Zhao, Jiahao and Wang, Yaonan and Wu, Zijie and Feng, Mingtao and Ding, Renjie and Zhang, Hui},
  journal={IEEE Transactions on Industrial Informatics},
  year={2026},
  publisher={IEEE}
}

@article{liang2026double,
  title={Double Diffusion Policy for Robust Robot Learning via Human Guidance},
  author={Liang, Weixiang and Gong, Ying and Li, Xiongyi and Liu, Yinlong and Yang, Zhi-Xin},
  journal={IEEE Transactions on Industrial Informatics},
  year={2026},
  publisher={IEEE}
}

@article{fang2025flow,
  title={Flow policy: Generalizable visuomotor policy learning via flow matching},
  author={Fang, Yu and Zhang, Xuehe and Cheng, Haoshu and Zang, Xizhe and Song, Rui and Zhao, Jie},
  journal={IEEE/ASME Transactions on Mechatronics},
  volume={31},
  number={1},
  pages={140--150},
  year={2025},
  publisher={IEEE}
}

@article{yan2025maniflow,
  author       = {Ge Yan and
                  Jiyue Zhu and
                  Yuquan Deng and
                  Shiqi Yang and
                  Ri{-}Zhao Qiu and
                  Xuxin Cheng and
                  Marius Memmel and
                  Ranjay Krishna and
                  Ankit Goyal and
                  Xiaolong Wang and
                  Dieter Fox},
  title        = {ManiFlow: {A} General Robot Manipulation Policy via Consistency Flow
                  Training},
  journal      = {CoRR},
  volume       = {abs/2509.01819},
  year         = {2025}
}

@article{zhong2026efficient,
  title={Efficient One-Step Human Motion Prediction Under Occlusions via Constrained MeanFlow},
  author={Zhong, Ruirui and Hu, Bingtao and Wang, Xi Vincent and Song, Xiuju and Feng, Yixiong and Zhou, Mengchu and Wang, Lihui and Tan, Jianrong},
  journal={IEEE Transactions on Industrial Informatics},
  year={2026},
  publisher={IEEE}
}

@article{liu2026mspgrasp,
  title={MSP-Grasp: Multiscale Perceptual Framework for 6-DoF Grasping in Cluttered Environments},
  author={Liu, Zhan and Wang, Ziwei and Chen, Lei and Zhou, Jie and Lu, Jiwen},
  journal={IEEE Transactions on Industrial Informatics},
  year={2026},
  publisher={IEEE}
}

@inproceedings{shridhar2023perceiver,
  title={Perceiver-actor: A multi-task transformer for robotic manipulation},
  author={Shridhar, Mohit and Manuelli, Lucas and Fox, Dieter},
  booktitle={Conference on Robot Learning},
  pages={785--799},
  year={2023},
  organization={PMLR}
}

@inproceedings{goyal2023rvt,
  title={Rvt: Robotic view transformer for 3d object manipulation},
  author={Goyal, Ankit and Xu, Jie and Guo, Yijie and Blukis, Valts and Chao, Yu-Wei and Fox, Dieter},
  booktitle={Conference on Robot Learning},
  pages={694--710},
  year={2023},
  organization={PMLR}
}

@inproceedings{gervet2023act3d,
  author       = {Th{\'{e}}ophile Gervet and
                  Zhou Xian and
                  Nikolaos Gkanatsios and
                  Katerina Fragkiadaki},
  title        = {Act3D: 3D Feature Field Transformers for Multi-Task Robotic Manipulation},
  booktitle    = {CoRL},
  series       = {Proceedings of Machine Learning Research},
  volume       = {229},
  pages        = {3949--3965},
  publisher    = {{PMLR}},
  year         = {2023}
}

@inproceedings{wang2024rise,
  title={Rise: 3d perception makes real-world robot imitation simple and effective},
  author={Wang, Chenxi and Fang, Hongjie and Fang, Hao-Shu and Lu, Cewu},
  booktitle={2024 IEEE/RSJ International Conference on Intelligent Robots and Systems (IROS)},
  pages={2870--2877},
  year={2024},
  organization={IEEE}
}

@inproceedings{qi2017pointnet,
  author       = {Charles Ruizhongtai Qi and
                  Hao Su and
                  Kaichun Mo and
                  Leonidas J. Guibas},
  title        = {PointNet: Deep Learning on Point Sets for 3D Classification and Segmentation},
  booktitle    = {{CVPR}},
  pages        = {77--85},
  publisher    = {{IEEE} Computer Society},
  year         = {2017}
}

@article{qi2017pointnet++,
  title={Pointnet++: Deep hierarchical feature learning on point sets in a metric space},
  author={Qi, Charles Ruizhongtai and Yi, Li and Su, Hao and Guibas, Leonidas},
  journal={Advances in neural information processing systems},
  volume={30},
  year={2017}
}

@article{ke2026surface,
  title={Surface Constraint Policy for Learning Surface-Constrained and Dynamically Feasible Robot Skills},
  author={Ke, Shuai and Zhang, Jiexin and Zhao, Huan and Wei, Zhiao and Guo, Yikun and Pan, Jie and Ding, Han},
  journal={IEEE Transactions on Industrial Informatics},
  year={2026},
  publisher={IEEE}
}

@article{xu2026diffog,
  author       = {Zhengtong Xu and
                  Zichen Miao and
                  Qiang Qiu and
                  Zhe Zhang and
                  Yu She},
  title        = {DiffOG: Differentiable Policy Trajectory Optimization With Generalizability},
  journal      = {{IEEE} Trans. Robotics},
  volume       = {42},
  pages        = {204--223},
  year         = {2026}
}

@article{xu2024leto,
  title={LeTO: Learning constrained visuomotor policy with differentiable trajectory optimization},
  author={Xu, Zhengtong and She, Yu},
  journal={IEEE Transactions on Automation Science and Engineering},
  volume={22},
  pages={8567--8578},
  year={2024},
  publisher={IEEE}
}

@article{chen2026phaseconditioned,
  title={Phase-Conditioned Imitation Learning With Autonomous Failure Recovery for Robust Deformable Object Manipulation},
  author={Chen, Dayuan and Tang, Kai and Zhang, Yukuan and Kosuge, Kazuhiro and Hirata, Yasuhisa},
  journal={IEEE/ASME Transactions on Mechatronics},
  year={2026},
  publisher={IEEE}
}

@article{wei2026hierarchically,
  title={Hierarchically Aware Replay-Enhanced Diffusion Policy for Robot Continual Imitation Learning},
  author={Wei, Zhiao and Zhao, Huan and Zhang, Jiexin and Ke, Shuai and Guo, Yikun and Ding, Han},
  journal={IEEE Transactions on Industrial Informatics},
  year={2026},
  publisher={IEEE}
}

@article{loaiza2024deep,
  title={Deep generative models through the lens of the manifold hypothesis: A survey and new connections},
  author={Loaiza-Ganem, Gabriel and Ross, Brendan Leigh and Hosseinzadeh, Rasa and Caterini, Anthony L and Cresswell, Jesse C},
  journal={arXiv preprint arXiv:2404.02954},
  year={2024}
}

@inproceedings{schenck2017learning,
  title={Learning robotic manipulation of granular media},
  author={Schenck, Connor and Tompson, Jonathan and Levine, Sergey and Fox, Dieter},
  booktitle={Conference on Robot Learning},
  pages={239--248},
  year={2017},
  organization={PMLR}
}

@inproceedings{dasari2024dit,
  title={The ingredients for robotic diffusion transformers},
  author={Dasari, Sudeep and Mees, Oier and Zhao, Sebastian and Srirama, Mohan Kumar and Levine, Sergey},
  booktitle={2025 IEEE International Conference on Robotics and Automation (ICRA)},
  pages={15617--15625},
  year={2025},
  organization={IEEE}
}

@article{balasubramanian2011robust,
  title={A robust and sensitive metric for quantifying movement smoothness},
  author={Balasubramanian, Sivakumar and Melendez-Calderon, Alejandro and Burdet, Etienne},
  journal={IEEE transactions on biomedical engineering},
  volume={59},
  number={8},
  pages={2126--2136},
  year={2011},
  publisher={IEEE}
}
\vfill 
\end{document}